# Traceable LLM-based validation of statements in knowledge graphs

Daniel Adam[1][0009-0007-3330-0148] and Tomáš Kliegr [1*][0000-0002-7261-0380]

[1] Prague University of Economics and Business, Faculty of Informatics and Statistics, Department of Information and Knowledge Engineering, Prague, Czech Republic
* Corresponding author
`tomas.kliegr@vse.cz`

**Abstract.** This article presents a method for verifying RDF triples using LLMs, with an emphasis on providing traceable arguments. Because the LLMs cannot currently reliably identify the origin of the information used to construct the response to the user prompt, our approach is to avoid using internal LLM factual knowledge altogether. Instead, verified RDF statements are compared to chunks of external documents retrieved through a web search or Wikipedia. To assess the possible application of this retrieval augmented generation (RAG) workflow on biosciences content, we evaluated 1,719 positive statements from the BioRED dataset and the same number of newly generated negative statements. The resulting precision is 88 %, and recall is 44 %. This indicates that the method requires human oversight. We also evaluated the method on the SNLI dataset which allowed us to compare our approach with models specifically tuned for the natural language inference task. We demonstrate the method on Wikidata, where a SPARQL query is used to automatically retrieve statements needing verification. Overall, the results suggest that LLMs could be used for large-scale verification of statements in KGs, a task previously unfeasible due to human annotation costs.

## 1. Introduction

Recent research has shown that many natural language processing (NLP) tasks, which previously required dedicated algorithmic approaches and model training, such as relation extraction or named entity recognition, can be performed by LLMs in a zero-shot manner. In this article, we investigate using zero-shot LLMs for validating statements in knowledge graphs. This task differs from other NLP tasks in that it emphasises the correctness of the output and particularly its traceability: an integral part of the validation process is the provision of information on why the verified statement is true, providing a reference to a specific part of a document addressing the topic in question. Despite the advances in LLM technology, between 50 % to 90 % of LLM responses are not fully supported by the sources provided (Wu et al, 2024). This article presents an intrinsically traceable approach to verification, which is based on avoiding using internal LLM factual knowledge and instead uses the LLM to find justification for the statement in the supplied text snippet. This method directly generates references, eliminating the error-prone phase of attribution of the response of the LLM to the training data (Lee et al, 2024).

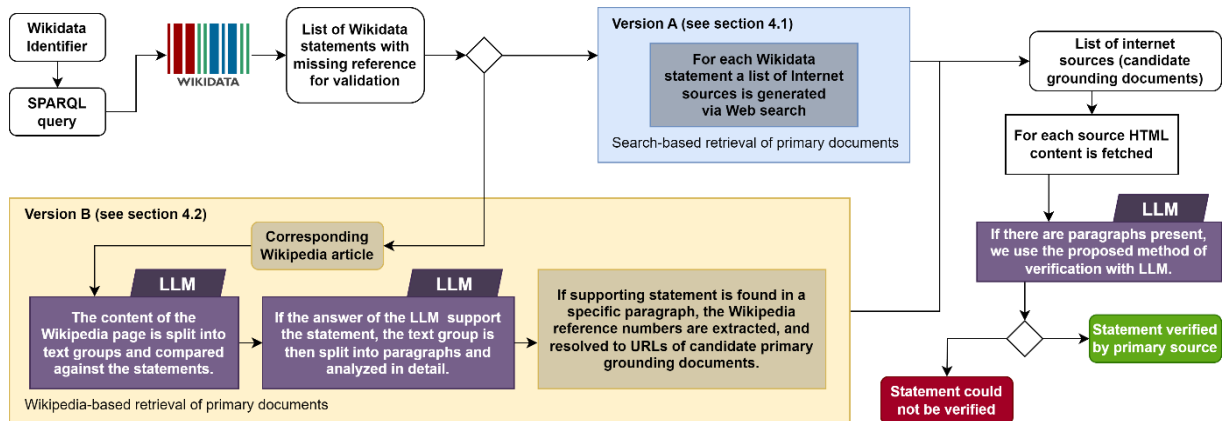

*Figure 1. Graphical abstract describing the proposed method for Wikidata statement verification*.

Figure 1 presents the proposed end-to-end LLM-based workflow for validating statements in Wikidata. The method has two variants: Version A retrieves grounding documents used for verification through web search, and Version B retrieves them from Wikipedia references. In Figure 2, we show the prompts used to validate a specific triple given a snippet from the input document against which the statement is to be verified. Based on the information in the text snippet, the LLM answered correctly and also provided a justification. For comparison, a naïve approach that would directly ask the same LLM (Llama 3 70B Instruct) to verify the statement based on its pretrained knowledge (and without providing the external snippet) would result in an incorrect negative answer.[1] To more rigorously evaluate the precision and recall of the proposed prompting methodology, we report on an evaluation performed on several thousand statements.

---

[USER INPUT] Can the given RDF be inferred from the given snippet? RDF for verification: ["Václav Havel" - "award received" - "Concordia Prize"]. Snippet to verify from: "*Awards Received: honorary doctor of the Université libre de Bruxelles | Honorary doctor of the University of Liège | Collar of the Order of the Cross of Terra Mariana | Order of the Three Stars, 1st Class | Evelyn F. Burkey Award | participant in the resistance and resistance against communism |....*" Please, choose the correct option based on your answer![a]

a) The RDF statement can be directly verified from the snippet. The snippet contains direct proof. b) The snippet contains some indications of the truthfulness of the RDF.
c) The RDF statement definitely cannot be inferred from the snippet.

[LLM OUTPUT] The correct answer is: a) The RDF statement can be directly verified from the snippet. The snippet contains direct proof. The snippet explicitly mentions \"Concordia Prize\" in the list of awards received by Václav Havel, which directly verifies the RDF statement.

[a]https://www.ask-oracle.com/birth-chart/vaclav-havel/, retrieved on August 29th.

---

*Figure 2. Example of a prompt-based verification based on externally provided candidate grounding text*

This article is structured as follows. In Section 2, we review related work, describing the differences from previous LLM-based approaches. We also review algorithmic steps which we incorporated from prior research on designing LLM-based NLP pipelines. Section 3 presents our prompt construction method. Section 4 describes in detail the two versions of the Wikidata verification workflow. Section 5 presents quantitative validation on the BioRED-Verify dataset. Section 6 presents quantitative validation on the SNLI dataset. In the last section, we interpret our results and suggest future work. The data, code, and detailed results for this research are openly available on our GitHub page.[2]

---

[1] Detailed outputs are available in the supplementary material.

## 2. Related work

Prior research on verifying RDF statements has so far largely relied on human annotators and crowdsourcing (Kontokostas et al., 2013; Acosta et al., 2013), which is a costly process that can be applied only to a small fraction of triples in need of verification. A detailed review of these approaches was done, e.g., by Xue and Zou, 2022. In the following, we will focus on the new generation of approaches that use LLMs, which have the potential to be more resource-efficient and allow for large-scale deployment.

The closest work to ours is the MiniCheck system (Tang et al., 2024), which is a trained sentence-level LLM-based fact-checker trained on synthetically generated data with GPT-4. Our approach is similar in that we also strive not only to verify a specific statement but also to provide evidence for that claim by linking the statement to a grounding document. However, in Tang et al., 2024, the grounding is a post-hoc process separate from the actual fact check. This, together with the need to pre-train a specific model, results in lower transparency and increases the complexity of the process. On the other hand, our approach uses an open-weight LLM without any pretraining (zero-shot). In our approach, the LLM is used as a tool to compare information in the grounding document with the verified statement. For this, we do not need pretraining, in fact, a key element in our design is to instruct the LLM not to use any of its internal subject matter expertise. Also, the MiniCheck system was designed as a generic fact-checking tool. In contrast, our approach was specifically designed for RDF statements, and, as such, it covers the identification of knowledge graph statements in need of validation. It also includes a Wikidata-specific approach to the retrieval of grounding documents, which bypasses the need to use internet search. On the other hand, we do not need to address the decomposition of input text into facts like generic fact-checkers do since individual RDF statements already correspond to atomic facts.

Wei et al., 2023 investigated the performance of zero-shot LLMs for several information extraction tasks relevant to our goal to perform zero-shot LLM-based verification of RDF statements. The evaluation on multiple datasets has shown competitive results to full-shot systems. This was achieved by dividing the task into several simpler subtasks corresponding to individual prompts, which we also performed. Instrumental to the result is a suitable prompt structure; specifically, the authors obtained better results with chat-like prompts (containing questions rather than a list of tasks), which we also adopt in our approach.

A comprehensive evaluation of ChatGPT on the information extraction task performed by Han et al., 2023 identifies additional recommendations for improving LLM-based information extraction results. For our purposes, the most significant is the observation that irrelevant contexts before and after the core text fragment can sometimes lead to a decrease in accuracy of up to 48 %. In our LLM chaining workflow, we reflect this by breaking down the grounding documents into paragraphs, which are independently compared against the verified statement.

Relevant is also the recent study of Efeoglu and Paschke, 2024, who evaluated fine-tuned LLMs within Retrieval Augmented Generation (RAG) workflows with a focus on sentence-level relations. While our approach could also be viewed as having an RAG component, an important differentiator from most RAG workflows is that we do not aim to blend the retrieved external text with LLM's trained knowledge.

---

[2] https://github.com/danieladam2001/LLM-based-validation

Another finding made by Han et al., 2023 is that ChatGPT (version not given) does not often understand the directions of relations between two concepts well, especially if these are asymmetrical relations. In an experiment they swapped the order of subject and predicate concepts and observed that only in 30 % of the situations the model recognized the inaccurate direction. We reflected this finding when designing our evaluation dataset, into which we included asymmetric relations.

An important element of our approach is the ability to generate traces, or explanations of the verifications. The verification result is technically a new edge, which links the verified statement with an external document that provides a reference and a justification. The verification task can thus be viewed as a particular instance of a link prediction task for which there is a growing interest in generating explanations. Unlike existing works that insert links between nodes already present in the graph and thus use information in the graph to generate the explanations (d'Amato et al, 2021), we generate the explanation from the external content and using an LLM.

A possible practical application of our work is a large-scale verification of facts in Wikidata. Although LLM-based verification may be more resource-effective than human-based verification, a cost-benefit analysis should be conducted to weigh the financial and environmental costs against the number of incorrect statements that could be identified by such a process. To this end, Santos et al., 2024 investigate the trustworthiness of Wikidata statements. They mainly focus on statements that share the same subject and predicate but connect to a different object. While in some cases, such statements are not errors (for example, one person can be in the *award received* relationship with multiple awards), these statements could be shortlisted for verification, especially if they miss some type of required qualifier. Santos et al., 2024 established the prevalence of statements with multiple values that have one or more mandatory required qualifier constraint violations at 0.24 %, amounting to over a million such statements in need of verification in Wikidata. There were other previous attempts at automatically identifying errors through machine learning, such as the work of Rico et al, 2018. However, this and similar approaches were able to spot errors typically on the schema or metadata level, while our work aims to verify the factual correctness of arbitrary statements using external knowledge.

## 3. Proposed method: Traceable LLM-based claim verification

This section describes the core of the proposed method: the verification of triples with LLMs *against a given snippet of text*. Here, we focus on the prompts used to reflect the paradigm that the model is asked to only use the information explicitly mentioned in the given snippet to *suppress the use of the model's internal knowledge*. In Section 4, we describe how this method is used to verify Wikidata statements, including the lookup of internet sources for Wikidata entries with missing references or parsing of Wikipedia entries for primary references.

### 3.1. Prompt generation

The prompt requires four text inputs. The subject, predicate, and object of the RDF statements and the text snippet from the grounding document against which the statement is verified. The identification of RDF statements for verification can be performed in several ways, as discussed in Section 4.

### 3.1.1. Text chunking

When transforming input text into text snippets, it is necessary to account for the maximum input token limit of each model. While the closed-source state-of-the-art (as of writing) models developed by OpenAI, such as GPT-4o mini and GPT-4o Turbo can receive up to 128.000 tokens[3], which corresponds to approximately 96.000 words, some open-source models, such as Llama 2 70B by Meta can receive only up to 4.000 tokens[4] or 3.000 words. We used the following chunking methods.

1. Division by paragraphs.
2. Filling the input token limit.

The advantage of the first method is that it is the most natural, while the second method results in increased efficiency as fewer total requests are sent to the LLM API. In Use Case 1 (section 4.1), where we process documents retrieved by web search, we used the division by paragraphs, and in Use Case 2, which verifies statements using the corresponding Wikipedia page, we used a combination of both approaches. The Wikipedia article is first divided into chunks of text. Each chunk comprises paragraphs that together do not exceed the threshold of 10,000 characters. If a supportive statement is found in the chunk of text, each paragraph is examined separately.

### 3.1.2. Prompt structure

The structure of the prompt used is shown in Figure 3. The prompt was designed so that the output of the LLM can be used for binary classification of the statement, either as confirmed by the passed snippet or as not being confirmed. There are three answer options, but only the first one (A), is considered as a positive response. Options B and C are considered as negative. This choice favours precision. If the unsure option B is excluded or interpreted as positive in downstream processing, the recall will increase, and precision will decrease.

---

Can the given RDF be inferred from the given snippet? RDF for verification: ["**SUBJECT**" - "**PREDICATE**" - "**OBJECT**"]. Snippet to verify from: "**EXTERNAL_TEXT_SNIPPET**" Please, choose the correct option based on your answer!

a) The RDF statement can be directly verified from the snippet. The snippet contains direct proof.
b) The snippet contains some indications of the truthfulness of the RDF.
c) The RDF statement definitely cannot be inferred from the snippet.

---

*Figure 3. Prompt structure used in the experiment: a specific example is presented in the introduction*

An example of an LLM output to the prompt is shown in Figure 4.

---

[LLM OUTPUT] The correct answer is: a) The RDF statement can be directly verified from the snippet. The snippet contains direct proof. The snippet explicitly mentions \"Previous winners of the award... include former Czech president, Václav Havel\", which directly verifies the RDF statement [\"Václav Havel\" - \"award received\" - \"Gottlieb Duttweiler Prize\"].

---

*Figure 4. Example response to the prompt shown in Figure 3*

As shown above, the model successfully chose one of the options and mentioned it explicitly in its answer. In our workflow, the answer given by the LLM is extracted from the response with a regular expression. The reasoning which led the model to choose one of the answers with respect to the snippet is, in approximately 99 % of cases, also included in the LLM answer. In the next section, we show how this reasoning is programmatically included in the explanation traces for the verification.

---

[3] https://www.scriptbyai.com/token-limit-openai-chatgpt/
[4] https://www.databricks.com/blog/llama2-inference

## 4. Application in knowledge graphs: Verifying data in Wikidata

Whether the content was added manually or using software, a small portion of the data in knowledge graphs is erroneous (Xue and Zou, 2022). An automated verifying system could help ensure high-quality data in knowledge bases in a cost-efficient and timely manner.

Wikidata belongs to the Wikimedia group of projects, including also Wikipedia and Wiktionary. Wikidata is a central publicly available knowledge base that serves as a foundation for other Wikimedia projects; thus, the integrity of its information does not affect only its direct users but also the visitors of Wikipedia. Wikidata contains approximately 1.5 billion item statements, which are represented as RDF triples (Wikipedia contributors, 2024).

In the following, we will present two use cases to verify statements in Wikidata. The two use cases differ in several ways to demonstrate the versatility of the approach, as shown in Table 1.

| Evaluation | Use case 1 | Use case 2 | BioRED-Verify dataset |
|---|---|---|---|
| Verified statements | Václav Havel is awarded Gottlieb Duttweiler Prize | Five statements with subject Bioluminiscence | 3,438 statements (1719 true positive and 1719 true negative) |
| Domain | General knowledge | Biochemistry | Biomedicine |
| Identification of verified statements | SPARQL query identifies statements with mandatory but missing reference | User-designated | Positive statements from BioRED dataset, negative statements generated according to Bordes et al., 2013. |
| Identification of candidate grounding documents | Web search | Through references in Wikipedia articles linked to the Wikidata entry | Fragments of research papers (set of 600 PubMed abstracts) included in the BioRED dataset |
| LLM used | LLama 3 70B Instruct | ChatGPT 4.0 -1106-preview | LLama 3 70B Instruct |

*Table 1. Overview of use cases and experimental validation.*

### 4.1. Use case 1: Using web search to retrieve candidate grounding documents

In this use case, a web search is used to identify candidate grounding documents, while the second use case covered in Section 4.2 takes advantage of the connection between Wikidata entries and Wikipedia to identify links to grounding documents from Wikipedia pages. Note that an overall workflow of both use cases was included in the graphical abstract in Figure 1.

#### 4.1.1. Identifying unsourced triples through SPARQL query

While RDF triples correspond to atomic facts, a typical pattern is that there is a core statement and then multiple supportive statements that elaborate on the core statement. For example, a core statement that a person was awarded a certain award can be accompanied by a supportive statement (qualifier) that indicates the year the award was received. Another important type of supporting statement is a citation of a grounding document that provides a source for the core claim. Some statements with missing citations are flagged by the Wikidata interface, as shown in Figure 5.

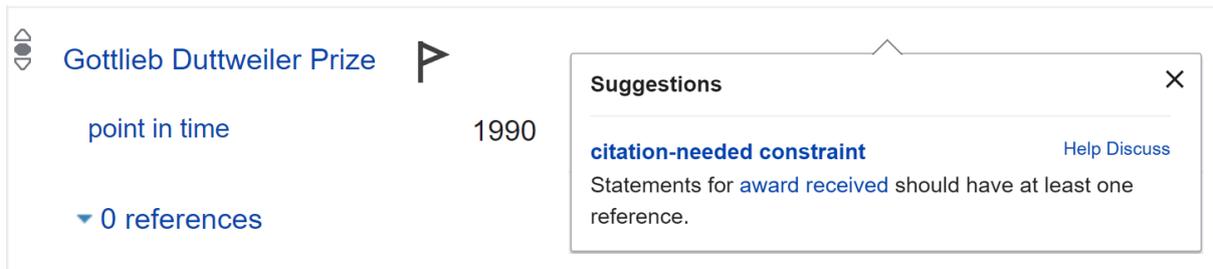

*Figure 5. A screenshot of a Wikidata statement about Václav Havel violating the citation constraint (Q36233, 2024)*

In this use case, we focus the verification on Wikidata statements, as exemplified above, which are those that do not meet the condition that a claim must have an assigned source (*reference*). Such statements can be retrieved from Wikidata SPARQL endpoint using the subject identifier (for example, Q36233 for Václav Havel) and SPARQL query, which returns a list of statements about the chosen subject that are missing a reference.

For some statement types, a citation is more important than for others, which is also differentiated in Wikidata: some predicates are associated with a constraint that mandates that a reference must be included when the predicate is used. Using a second SPARQL query, we select only those statements that have a constraint to contain a reference. If used with the subject `Václav Havel`, we get a list of 53 unsourced statements, a sample of which is shown in Table 2.

| **Subject** | **Predicate** | **Object** |
|---|---|---|
| Václav Havel | award received | Gottlieb Duttweiler Prize |
| Václav Havel | award received | Philadelphia Liberty Medal |
| Václav Havel | award received | Quadriga |
| Václav Havel | award received | Point Alpha Prize |
| Václav Havel | award received | Concordia Prize |
| Václav Havel | award received | St. George's Order of Victory |

*Table 2. Sample of statements for subject with Václav Havel (identifier Q36233) that were identified using a SPARQL query to have missing references although a reference is mandatory for the given predicate.*

*4.1.2. Finding candidate grounding documents using a web search engine*

Next, we take each statement and use it as an input for automated Google search. We only use the subject, predicate and object separated by space and add a parameter "-wikipedia" to exclude results containing Wikipedia (Wikipedia-based verification is a separate mode of operation described in Section 4.2). We also limit the search results to the top five hits. Finally, we use a Python library Requests[5] to retrieve the HTML content of the retrieved resources.

When applied for the first statement ("Václav Havel", "award received", "Gottlieb Duttweiler Prize") we obtained a list of five URLs, with the first three being web pages of different winners of the Gottlieb Duttweiler Prize and the last two being annual reports (in pdf) from the Vaclav Havel foundation. Detailed results are available from the GitHub repository referenced from the Data availability section included at the of the manuscript. Since non-HTML content such as PDF is currently unsupported by our implementation, only the first three URLs enter into the LLM-based verification.

---

[5] https://pypi.org/project/requests/

*4.1.3. LLM-based verification*

If the content of the retrieved document is available, we go through all paragraph tags and if their content is valid (at least 100 characters long), we use them as individual inputs to the LLM-based verification. Subsequently, we invoke the prompt from Figure 3 on the individual paragraphs. If the LLM returns a positive answer, we end the loop and save the current paragraph that yielded the confirmation for documentation purposes. In this case, the remaining document content is not processed to conserve resources. When a match is found, or none was found but all paragraphs of the current document were processed, we proceed to the next document.

For our sample statement, the verification process was executed against all five retrieved documents described in Section 4.1.2. The LLM (LLama 3 70B Instruct) found a supporting paragraph in documents number 1 and 3, both mentioning the award to Václav Havel when announcing new award winners. The verification result for the example statement, according to the first document, is included in Figure 6.

---

**Verification trace for "<Václav Havel> <award received> <Gottlieb Duttweiler Prize>"**

The LLM (Llama 3) found support for the verified statement in the following paragraph of the first document[a]: "*IBM believes the promise of technology is to empower people to do good. We are honored for this belief to have been reinforced last night, when we received such a renowned award.* **An award which has also been given to the likes of Václav Havel**, *Czech politician, writer and human rights activist and Tim Berners-Lee, the inventor of the World Wide Web, for their outstanding contributions to the well-being of the wider community and to cultural, social or economic environments. Now, for the first time in history,* **the Gottlieb Duttweiler Institute** *has made the decision to honor a technology.*"

The output of the LLM provided this justification: "**The correct answer is a)** *The RDF statement can be directly verified from the snippet. The snippet* **contains direct proof**. *The snippet explicitly mentions that the award has been given to Václav Havel, and the award is from the Gottlieb Duttweiler Institute, which matches the RDF statement [\"Václav Havel\" - \"award received\" - \"Gottlieb Duttweiler Prize\"]*."

[a]https://www.linkedin.com/pulse/gottlieb-duttweiler-prize-2019-goes-watson-matthias-hartman, retrieved on August 29th.

---

*Figure 6 Verification results for the statement "<Václav Havel> <award received> <Gottlieb Duttweiler Prize>".*

We verified the reasoning in Figure 6 as well as the second positive result and found no problems. Both responses were correct.

For the initial Wikidata statement, which missed a reference even though its predicate was mandated to have one, this process found two grounding documents, providing a traceable justification of the relation between the statement and each of the documents. The entire process of this statement, including the search and retrieval of all five URLs and their analysis, took about 15 seconds. Detailed results are available from the GitHub repository referenced from the Data availability section included at the of the manuscript.

### 4.2. Verifying unsourced Wikidata statements by a corresponding Wikipedia page

This use case is aimed at the verification of specific Wikidata statements from the biological domain. We manually chose five statements about the subject "Bioluminescence" (Q179924) shown in Table 3. We used ChatGPT-1106-preview LLM for this experiment.

| Subject | Predicate | Object |
|---|---|---|
| Bioluminescence | different from | iridescence |
| Bioluminescence | different from | biofluorescence |
| Bioluminescence | different from | phosphorescence |
| Bioluminescence | different from | fluorescence |
| Bioluminescence | subclass of | cell metabolism |

*Table 3. List of Wikidata statements chosen for verification in Use Case 2*

#### 4.2.1. Mapping Wikidata entries to Wikipedia

Unlike in Use Case 1 (Section 4.1.2) where the grounding documents are obtained directly via a web search, in this use case, we utilize the correspondence between the Wikidata entries and Wikipedia articles. The name of the subject, which was chosen as a parameter (Bioluminescence), is used in a Wikidata API call, which returns the URL of the corresponding Wikipedia page.

#### 4.2.2. Identifying grounding documents from Wikipedia references

Once the Wikipedia article corresponding to the subject of the verified statements has been established, its content is fetched and split first into larger text groups. Given the typical longer length of Wikipedia articles and the relatively lower likelihood that the subject will even be mentioned in a specific paragraph, we split the documents into chunks of several paragraphs each (observing the maximum of 10,000 characters) to increase the efficiency of the processing. In the first phase, the prompt in Figure 3 was used on these larger chunks. If there was a positive result, the chunk was broken down into paragraphs and the process repeated. A positive match was thus associated with a precise part (a paragraph) of the Wikipedia article. However, since Wikipedia is not generally considered an authoritative source on its own, the location of the grounding text in the Wikipedia article was used as input for subsequent processing aimed at finding the primary grounding document. To do this, we extracted the reference numbers that were associated with the identified paragraph in the Wikipedia article and extracted the corresponding citations from the References section of the article.

*4.2.3. LLM validation against primary grounding document*

After the process obtained one or more references from the Wikipedia article, these are attempted to be downloaded. If the resource is not available, we attempt to obtain its archived copy from the Web Archive (archive.org).

Once we received the content either directly or via the Web Archive, we performed the proposed method described in Section 3 – we split the content into paragraphs and iteratively applied the prompt in Figure 3 to verify the RDF statement against each of the paragraphs.

An example automatically generated summary of the verification process is shown in Table 4.

| Attribute | Value |
| --- | --- |
| LLM used: | gpt-4-1106-preview |
| Date of testing: | 2024-03-02 |
| Time of testing: | 13:27:25 |
| Duration (minutes): | 24 minutes |
| Duration (seconds): | 39 seconds |
| Subject ID: | 179924 |
| Subject name: | Bioluminescence |
| Subject URL: | http://www.wikidata.org/entity/Q179924 |
| Subject Permalink: | https://wikidata.org/w/index.php?title=Q179924&oldid=2073869683 |
| Wikipedia URL: | https://en.wikipedia.org/wiki/Bioluminescence |
| Wikipedia Permalink: | https://en.wikipedia.org/w/index.php?title=Bioluminescence&oldid=1206514418 |
| Wikidata Endpoint: | https://query.wikidata.org/sparql |

*Table 4. Summary results for verification of the five statements about Bioluminiscence included in Table 3.*

A detailed verification result for one of the statements is included in Table 5. According to this, bioluminescence is different from biofluorescence, with this verification being traceable to a specific passage in a journal article referenced from Wikipedia.

| Attribute | Value |
|---|---|
| Content of paragraph received from the primary source | Phosphorus was thought to be the source of light in living creatures. Researchers now know that bioluminescence is accomplished through oxidation (the addition of oxygen) in an animal protein called luciferin. When a molecule of oxygen, either in a gaseous form or mixed in a liquid, and an enzyme known as luciferase combine with luciferin, the resulting new molecule is excited and gives off light. Unlike fuel combustion, there is no heat associated with luminescence. |
| Answer of the LLM | Yes, the RDF statement can be inferred from the given snippet. Reasoning: The snippet explains the biochemical process behind bioluminescence, which involves an animal protein called luciferin and the enzyme luciferase in the presence of oxygen to produce light without heat. It contrasts this with "fuel combustion", which is known to produce both light and heat, but what it implicitly suggests is that bioluminescence is a specific type of luminescence that is biochemically distinct. Phosphorescence, although not directly mentioned or described in the snippet, is also a type of luminescence, like bioluminescence; however, it fundamentally differs in its mechanism—where phosphorescence involves the delayed re-emission of light after absorption. Since the snippet points out the unique mechanism by which bioluminescence operates, without any reference to phosphorescence, it can be inferred that bioluminescence is indeed different from phosphorescence (and other forms of luminescence such as fluorescence), even though the distinctions between these processes are not explicitly discussed. Therefore, based on the information provided, it is implicitly understood that bioluminescence is a distinct phenomenon within the category of luminescence, making it different from phosphorescence. Hence, we can infer the "different from" relationship between "Bioluminescence" and "phosphorescence" from the snippet. |
| Match found in primary source? | Yes, according to the LLM the given paragraph from the primary source supports the RDF! |

Table 5. Part of an automatically generated report presenting a detailed comparison against a specific paragraph for the statement "bioluminescence is different from biofluorescence". The quoted text in the field "Content of paragraph received from the primary source" is sourced from Reshetiloff (2001). The content in the field Answer of the LLM was generated by "ChatGPT-4-1106-preview".

### 4.3. Reference implementations

Both use cases were implemented as generic Python applications, which are available from the Github repository referenced from the Data availability section included at the of the manuscript. Since the recursive search through Wikipedia articles generates multiple outputs, a reference implementation of the proposed workflow also supports automatic documentation following the W3C XML standard technologies. A complete process trace is saved to an XML output file conformant to an XSD document that validates this XML. This is accompanied by an XSL Transformation, which produces the HTML reports, samples of which are shown in Tables 4 and 5.

## 5. Evaluation on the BioRED dataset

The evaluation of the proposed method for verifying RDF triples was conducted on a new dataset derived from BioRED dataset (Luo et al., 2022), which covers five concept types (such as chemicals and diseases) and 8 relationship (predicate) types. Annotation was primarily done manually but supported by automatic tools such as TeamLat (Islamaj et al., 2020) and PubTator (Wei et al., 2019). As the BioRED dataset was originally designed for relation extraction and named entity recognition tasks, we had to perform several modifications and extensions to allow its use for the benchmarking of RDF statement verification methods. The summary of the resulting dataset, which we call BioRED-Verify, is included in Table 6, and the details are presented in the following subsection.

### 5.1. Dataset preparation: BioRED-Verify

From the original BioRED dataset, we used all concept types and their combinations (such as "GeneOrGeneProduct" or "ChemicalEntity"), except for the combination of "SequenceVariant" and "ChemicalEntity" which was excluded due to low occurrence (there were only 10 records in total for both "Positive_Correlation" and "Negative_Correlation").

Out of the eight relationship types contained in the original BioRED dataset, we chose the relations "Positive_Correlation" and "Negative_Correlation". The reason for this is that they are unambiguous and exclusive. For example, we considered the relation "Association", which is also part of the dataset, inappropriate for our purposes because it is a general relation that subsumes both positive and negative correlation. An overview of dataset size by relation is included in Table 6. The details on how the positive and negative examples were generated are in the following subsections.

| Relation type | Positive examples | Negative examples |
|---|---|---|
| Positive correlation | 1019 | 1019 |
| Negative correlation | 700 | 700 |

Table 6. Size of BioRED-Verify dataset

#### 5.1.1. Positive examples

As mentioned before, BioRED is a relation-extraction dataset which means its structure and format are unsuitable for a verification task. The dataset consists of articles with each article being associated with a list of entities and a list of relations between these entities that can be found in the article. The original use of the dataset was for information extraction methods to identify relations in the given text, which were then confirmed against the ground truth relations in the dataset. We essentially reversed this task by using the dataset to evaluate whether the associated text supports a given ground truth relation from BioRED. In this way, we obtained the positive examples.

#### 5.1.2. Negative examples

Since the BioRED dataset contained no negative examples, these had to be introduced. We decided to use the method of creating corrupted triples by replacing the tail entity of the triple (the object) with a different entity of the same type. This method is used in frequently referenced link prediction datasets such as WN18RR (Bordes et al., 2013), derived from WordNet, and FB15k-237 (Toutanova and Chen, 2015), derived from Freebase. Also, following the protocol for generating a filtered version of the dataset according to Bordes et al., 2013, we excluded any randomly created corrupted triples that corresponded with existing ground-truth triples.

## 5.2. Evaluation using Llama 3 70B

Using Llama 3 70B and the API service hosted by Replicate.com we evaluated the workflow proposed in Section 3 on the BioRED-Verify dataset.

| Concept 1 | Concept 2 | TP | TN | FP | FN | Precision | Recall |
|---|---|---|---|---|---|---|---|
| ChemicalEntity | DiseaseOrPhenotypicFeature | 176 | 382 | 14 | 220 | 92.63 % | 44.44 % |
| ChemicalEntity | GeneOrGeneProduct | 63 | 122 | 2 | 61 | 96.92 % | 50.81 % |
| SequenceVariant | DiseaseOrPhenotypicFeature | 201 | 173 | 65 | 37 | 75.56 % | 84.45 % |
| ChemicalEntity | ChemicalEntity | 27 | 50 | 1 | 24 | 96.43 % | 52.94 % |
| GeneOrGeneProduct | GeneOrGeneProduct | 97 | 160 | 7 | 70 | 93.27 % | 58.08 % |
| GeneOrGeneProduct | DiseaseOrPhenotypicFeature | 25 | 40 | 3 | 18 | 89.29 % | 58.14 % |
| **Average (micro)** | | **589** | **927** | **92** | **430** | **86.49 %** | **57.80 %** |

*Table 7. Evaluation of predicate "Positive_Correlation"; TP: True Positive, TN :True Negative, FP: False Positive, FN: False Negative*

We tested both relations separately. The results for the positive correlation relation are shown in Table 7. Most concept combinations resulted in a precision of around 90 % or higher. The only exception was the combination of "SequenceVariant" and "DiseaseOrPhenotypicFeature" with precision of 76 %. This shows that the concept type combination RDF can have a noticeable impact on the verification performance.

The values of recall were, in most cases, in the range between 50 % and 60 %. A possible explanation is the comparatively high difficulty of the dataset, which contains specialized knowledge. We also observed an inverse relationship between precision and recall. While the precision for the combination "SequenceVariant" and "DiseaseOrPhenotypicFeature" was the smallest, this combination had the highest recall of 84 %.

| Concept 1 | Concept 2 | TP | TN | FP | FN | Precision | Recall |
|---|---|---|---|---|---|---|---|
| ChemicalEntity | DiseaseOrPhenotypicFeature | 37 | 277 | 1 | 241 | 97.37 % | 13.31 % |
| ChemicalEntity | GeneOrGeneProduct | 77 | 174 | 10 | 107 | 88.51 % | 41.85 % |
| SequenceVariant | DiseaseOrPhenotypicFeature | 9 | 13 | 0 | 4 | 100.00 % | 69.23 % |
| ChemicalEntity | ChemicalEntity | 18 | 110 | 0 | 92 | 100.00 % | 16.36 % |
| GeneOrGeneProduct | GeneOrGeneProduct | 29 | 65 | 4 | 40 | 87.88 % | 42.03 % |
| GeneOrGeneProduct | DiseaseOrPhenotypicFeature | 5 | 46 | 0 | 41 | 100.00 % | 10.87 % |
| **Average (micro)** | | **175** | **685** | **15** | **525** | **92.11 %** | **25.00 %** |

*Table 8. Evaluation of predicate "Negative_Correlation"; TP: True Positive, TN :True Negative, FP: False Positive, FN: False Negative*

The results of the test on the statements with negative correlation are shown in Table 8. The average precision was slightly higher. Two of the concept variations had 100 % precision, even though the combination of "SequenceVariant" and "DiseaseOrPhenotypic Feature" only consisted of 13 records, which may not be considered representative enough. On the other hand, it resulted in a markedly lower recall.

The average precision (micro average) across both predicates is 87.7 %, and the average recall is 44.4 %. This corresponds to the F1 score of 59.0 % for our zero-shot LLM-based approach. For comparison, the authors of the BioRED dataset reached an almost identical F1 score of 58.9 % in the relation extraction task using the model PubMedBERT (Luo et al., 2022). Note that the latter work did not include the negative statements in BioRED-Verify, but on the other hand, it covered the full BioRED dataset, which consists of 6,503 relations in total (training, development, and test sets combined).

### 5.3. Error analysis and low recall

In the following, we separately analyze the false negative cases, false positive cases and the model performance in terms of recall.

**False positives:** Almost all cases in which the LLM answered falsely positively were united by the mistake of the LLM *not understanding the RDF data format*. Instead of treating the input RDF statement as one statement, the LLM was, in some cases, satisfied with only a part of the RDF statement being supported by the text.

**False negatives**: One important source of error was the LLM's inability to find the relationship between the subject and the predicate in the text, although this was present. We also noticed several cases in which the LLM did find the relationship but still answered incorrectly due to an *erroneous disambiguation of the subject*. This typically happened when the text (PubMed abstract) did not contain the subject as exactly in the RDF, but instead, it contained its slightly reworded version. This confused the LLM into thinking it was an entirely different subject.

**Low recall.** One of the main reasons is the difficulty of the dataset, which deals with specialized biosciences knowledge. Also, given that the inverse relationship between recall and precision is a widely recognized phenomenon (Cleverdon et al., 1972), we faced a choice regarding whether to put more emphasis on recall or precision – and we chose to favour precision. This was reflected by using only three options as possible answers in Figure 3. Furthermore, the middle (neutral) option could be assessed as either a positive answer or a negative answer. Based on our design option to favour precision, we interpreted a positive answer to the neutral option as evidence for the statement not being supported. However, if we wanted to increase the recall (at the expense of lower precision), we could also add the neutral option to the pool of positive answers, which would result in a decrease in the false negative answers (and an increase in false positive answers).

### 5.4. Comparison with a baseline (a pre-trained cross-encoder model)

To compare our method with a baseline, we evaluated the BioRED dataset using a pre-trained cross-encoder model, nli-deberta-base[6], which was trained on the SNLI and MultiSNLI datasets. This model is primarily suited for sentence-to-sentence comparison, which requires some adaptation of the dataset used. Since the BioRED dataset contains PubMed abstracts, we separated the abstracts into individual sentences and evaluated each sentence against the RDF separately. Finally, we had to reconcile the results in case there were multiple contradicting sentence-level results that had to be merged into one final judgment for the whole abstract. For this, we employed two alternative methods, which we call the strict and loose approach.

---

[6] https://huggingface.co/cross-encoder/nli-deberta-base

The *loose approach* marks each BioRED entry as an entailment if at least one of the separately processed sentences was evaluated as an entailment. This allows having one or more sentences marked as a contradiction. As a result, this approach has higher recall but lower precision.

The *strict approach* works in the same way as the loose approach, but in addition, forbids containing any sentences marked as a contradiction. Therefore, if one or more sentences are marked as an entailment and the rest are marked as a neutral statement, the whole entry is evaluated as an entailment. If all sentences are marked as an entailment except for one which is a contradiction the whole entry is considered a contradiction.

The results of both methods are shown in Tables 9 and 10.

|  | TP | TN | FP | FN | Precision | Recall |
|---|---|---|---|---|---|---|
| Positive Correlation | 12 | 1019 | 0 | 1007 | 100 % | 1.18 % |
| Negative Correlation | 0 | 700 | 0 | 700 | NaN | 0 % |

*Table 9. Results of the strict approach of nli-deberta-base on the BioRED dataset. TP: True Positive, TN:True Negative, FP: False Positive, FN: False Negative.*

|  | TP | TN | FP | FN | Precision | Recall |
|---|---|---|---|---|---|---|
| Positive Correlation | 49 | 1012 | 7 | 970 | 87.5 % | 4.81 % |
| Negative Correlation | 55 | 695 | 5 | 645 | 91.67 % | 7.86 % |

*Table 10. Results of the loose approach of nli-deberta-base on the BioRED dataset. P: True Positive, TN:True Negative, FP: False Positive, FN: False Negative.*

Even though the nli-deberta-base model was not trained on a bio-scientific corpus but rather on a dataset containing general topics, it demonstrates high precision. Nevertheless, the F1 scores, which also reflect the value of recall, are both for the loose approach (F1 score of 11 %) and strict approach (F1 score of 1 %) lower than the values obtained for our method, as reported in Section 5.2.

### 5.5. Replicability

The model used for the verification task was Llama 3 70 B hosted at Replicate.com. To make the results replicable, we used a random seed and fixed hyperparameters as shown in Table 11. The dataset used for the evaluation is publicly available at https://ftp.ncbi.nlm.nih.gov/pub/lu/BioRED/. Llama 3 is an open LLM available at https://llama.meta.com/. The raw LLM responses used for evaluation are available at: https://github.com/danieladam2001/LLM-based-validation.

| Parameter | Value |
|---|---|
| model | meta/meta-llama-3-70b-instruct |
| seed | 42 |
| top_p | 0.9 |
| temperature | 0.1 |
| system_prompt | You are a helpful assistant. Work only with the text given to you. |
| max_new_tokens | 500 |
| min_new_tokens | -1 |

*Table 11. Parameters of the large language model for BioRED evaluation*

## 6. Evaluation on the SNLI dataset

To obtain additional insight into the performance of our LLM-based statement validation methodology, we evaluated three versions of the Llama LLM on the Natural language inference (NLI) task. NLI is an independent task in the field of NLP but shares many similarities with information extraction and the previously described version of our method for validating RDF statements. NLI is a process of assessing whether a hypothesis is an entailment, neutral, or a contradiction in the context of a premise.

To evaluate the performance on the NLI task, we used the SNLI dataset (Bowman et al., 2015), which contains 570, 000 records (split into a training set of 550,000 records, a development set of 10,000 records and a test set of 10,000 records). The dataset is not in the RDF format. Instead, it consists of sentence pairs, where one sentence is a premise and the other sentence is a hypothesis.

By slightly modifying the prompt, we replaced the original RDF input (as used with the BioRED dataset) with an input in plain sentence format, representing a hypothesis in the SNLI dataset. The text that served as the potential grounding document in the BioRED prompts represented the premise from the SNLI dataset.

While our LLM-based method does not need any training data, we used one record from the training set for each class (entailment, neutral, contradiction) to generate examples that were included in the prompts. Out of the 550,000 train set records, we used three randomly chosen training records as examples, the remaining training records were not used.

The evaluation was conducted on 1,000 entries that were randomly selected from the SNLI test set.

*6.1. Results*

The results are shown in Tables 12, 13, 14. The percentage in each cell was obtained by dividing the absolute number of occurrences (the value in parentheses) by 1,000, which is the number of records in the test set. The overall accuracy was 65.8 % for Llama 3 8B, 86.7 % for Llama 3 70B and 89.5 % for Llama 3.1 405B.

| llama-3-8b | Answered entailment | Answered neutral | Answered contradiction |
|---|---|---|---|
| Expected entailment | 32.6 % (326) | 1.5 % (15) | 0.4 % (4) |
| Expected neutral | 24.8 % (248) | 2.7 % (27) | 3.5 % (35) |
| Expected contradiction | 2.3 % (23) | 1.5 % (15) | 30.5 % (305) |

*Table 12. Llama 3 8B results. The total number of evaluated test set records was 1,000, and the sum of entries in the table is 998, as 2 entries were incorrectly parsed by the LLama 3 8B model.*

| llama-3-70b | Answered entailment | Answered neutral | Answered contradiction |
|---|---|---|---|
| Expected entailment | 32.6 % (326) | 0.9 % (9) | 1.1 % (11) |
| Expected neutral | 6.3 % (63) | 21.7 % (217) | 3 % (30) |
| Expected contradiction | 0.4 % (4) | 1.6 % (16) | 32.4 % (324) |

*Table 13. Llama 3 70B results. The total number of evaluated test set records was 1,000.*

| llama-3.1-405b | Answered entailment | Answered neutral | Answered contradiction |
|---|---|---|---|
| Expected entailment | 32.1 % (321) | 1.7 % (17) | 0.8 % (8) |
| Expected neutral | 3.8 % (38) | 26 % (260) | 1.2 % (12) |
| Expected contradiction | 0.2 % (2) | 2.8 % (28) | 31.4 % (314) |

*Table 14. Llama 3.1 405B results. The total number of evaluated test set records was 1,000.*

The obtained scores reveal an interesting insight into the effect of the number of parameters of the LLM model. While the expectation that the larger the model the better the performance was fulfilled, the difference between the 8B model and the 70B model is much more pronounced than the difference between the 70B and 405B models. The 70B appears to provide the best trade-off between model complexity and performance.

In terms of the performance in the three types of relationships, the entailment class was the least difficult for the models to assess. While Llama 3 70B performed particularly well at classifying the contradiction cases, Llama 3.1 405B surpassed both alternative models in classifying the neutral cases.

## 6.2. Error analysis

In this section, we show a detailed analysis of a falsely classified example for each Llama version. The raw LLM outputs are shown in Figures 7-9.

For Llama 3 8B, we observed the lowest accuracy. To illustrate the errors in model reasoning, we chose the following pair of the input premise "A man in suit waiting for the green light", and the hypothesis "A man waits at a red light". While the relationship between the two sentences is an entailment, the model falsely answered with a contradiction, providing the argument shown in Figure 7.

> [LLM OUTPUT] *Based on the given texts, I would say that the correct label is:c) contradiction. The reason is that the premise states that the man is waiting for a green light, but the hypothesis states that the man is waiting at a red light. This is a clear contradiction, as the man cannot be waiting for a green light if he is already at a red light.*

*Figure 7. An example of a false answer of Llama 3 8B model showing erroneous reasoning.*

The errors for Llama 3 70B version are characteristic of a very strict approach to assessing the similarity of subjects in a sentence. To illustrate this, we chose the following input pair wrongly processed by the model: "A woman is sitting in a chair, while a man is going through the sport utility vehicle" (premise) and "a girl is sitting" (hypothesis). While the annotators marked this entry as an entailment, the large language model answered with a contradiction, providing reasoning included in Figure 8.

> [LLM OUTPUT] *The correct answer is c) contradiction. The premise states "A woman is sitting in a chair", while the hypothesis states "a girl is sitting". The contradiction lies in the fact that "woman" and "girl" are not the same, so the hypothesis mentions something that would not be possible in the premise.*

*Figure 8. An analysis of a false answer of Llama 3 70B containing an erroneous disambiguation of the subject.*

Finally, we provide an example of a reasoning error characteristic of the largest model Llama 3.1 405B. We chose the input pair "Overly dramatic couple pose for a picture where an 'angry' man 'chokes' a woman who sticks out her tongue" (premise) and "A man chokes a woman" (hypothesis). While the entry was labeled as an entailment, Llama chose to assess this entry as a contradiction. As shown in Figure 9, it argued that the use of quotation marks suggests not taking the words literally, as opposed to the hypothesis where the quotation marks are missing.

> [LLM OUTPUT] *Based on the given texts, I would choose option c) contradiction. The premise states that the man "chokes" the woman, but it's in the context of an overly dramatic couple posing for a picture, and the quotation marks around "chokes" suggest that it's not meant to be taken literally. In contrast, the hypothesis states "A man chokes a woman" without any indication that it's not meant to be taken literally. Therefore, the hypothesis contradicts the premise because it implies a level of violence or aggression that is not present in the original context.*

Figure 9. An example of Llama 3.1 405B reasoning.

### 6.3. Comparison against other algorithms

The choice of the widely used SNLI dataset allows us to compare our LLM-based solution against a range of other algorithmic approaches. The papers-with-code repository contains benchmarking results for 98 models for this dataset.[7] The primary evaluation criterion, according to papers-with-code, is test set accuracy. Using this evaluation measure, our model based on Llama 3.1 405B would score in the top 20 % of solutions, surpassing, for example, the ESIM + ELMo Ensemble (Peters et al., 2018), which is based on a new type of deep contextualized word representation, or DEIM (Jiang et al., 2022), a deep encoding and interaction model for sentence matching. Llama 3.1 405B also achieved a slightly higher score than Parikh et al. (2016), who achieved an accuracy of 86.8 % using a decomposable attention model with intra-sentence attention.

Even though there are multiple models with higher accuracy, most of them have been trained on the entire training set. On the other hand, our approach almost did not use the training set - we only presented a single training example of a correct evaluation for each label to the LLM.

## 7. Discussion and conclusions

### 7.1. Use cases

We described two use cases using LLM-based verification for checking Wikidata content, which differ mainly in how the primary grounding documents were found. We found searching for internet sources directly via a web search API to be fast and efficient. Its disadvantage is that web search may return resources with varying credibility. The alternative option based on the mapping between Wikidata entries and Wikipedia articles is more resource-intensive and results in lower recall. Compared to the number of grounding candidates reachable by web search, Wikipedia articles contain only a smaller number of primary documents. However, the sources used in the References section of Wikipedia are subject to community-enforced quality guidelines[7], which may result in higher reliability of this method if it returns a positive result. Also, given the curated nature of the mapping between Wikidata and Wikipedia, the references obtained from Wikipedia are naturally disambiguated, which is not the case for retrieving documents from web search.

### 7.2. Key findings from quantitative evaluations

We performed evaluations of the proposed method on two datasets.

The first quantitative evaluation was performed on a modified version of the BioRED dataset containing 3,438 statements. The observed differences between the results for positive and negative correlation relations imply that the specific relation poses a distinct set of challenges for LLMs. From the application perspective, the comparatively high precision suggests that the intended workflow aimed at the use of LLMs to propose missing references is feasible. The average precision, as evaluated on the BioRED dataset, is 88 %. While this value may appear sufficient for some uses, in our view, the 12 % false positive rate still implies the need for human oversight. On the other hand, we consider the average recall at 44 % satisfactory, as typically, a given statement needs only one grounding document, while multiple grounding documents (or their confirming passages) are typically available, particularly if the web search candidate retrieval method is used.

---

[7] https://en.wikipedia.org/wiki/Wikipedia:Reliable_sources/Perennial_sources

As shown in the evaluation results, the performance varies significantly by the type of entities in the subject and object positions and by the type of the relation. Given that the data originated from a highly specialized biosciences domain, the predictive performance may be higher on common sense or general-knowledge topics. On the other hand, a real-world performance when used for Wikidata verification will also be affected by the success of the search for candidate grounding documents for which we proposed and implemented two variants.

The second quantitative evaluation was performed on 1,000 randomly chosen records from the SNLI dataset. We observed major differences between the results of the different Llama 3 versions, which implies that the right choice of the model can play an important role in the performance. Unlike the first quantitative evaluation, in this case, the LLM had to choose between three mutually exclusive classes, which made the decision-making process slightly more difficult.

### 7.3. Theoretical contributions

In this article, we described a novel zero-shot method for verifying statements in knowledge graphs. It differs from the previous work on LLM-based fact-checking in that it emphasizes utilizing the LLM as a tool to interpret and compare possible grounding documents with verified statements rather than relying on any domain-specific knowledge the LLM may have. The code for the proposed method is freely available, and the quantitative evaluation is reproducible, as it was carried out on openly available datasets using an open-weight LLM. The average precision at 88 % obtained on the BioRED-Verify dataset was unexpectedly high, considering we used a generic Llama 3 model with no finetuning. In one of the use cases, we have shown that the workflow can also be used with a closed LLM, potentially resulting in even better performance.

### 7.4. Practical implications

The proposed workflow with accompanying source code can foster large-scale, automated verification of knowledge graph statements. We also created and made available a new dataset, BioRED-Verify, to evaluate the performance of knowledge graph verification methods. While our use cases focused on widely known Wikidata, we foresee the main application in verifying statements in specialized biological and biochemical knowledge graphs, such as KG-Microbe (Santangelo et al., 2025) or Monarch (Shefchek et al, 2020). Our proposed workflow could be integrated into knowledge graph ingestion and maintenance frameworks such as KG-Hub (Caufield, 2023).

### 7.5. Limitations and future work

Our motivation for the approach taken in this article was an intuition that an LLM's internal representation of facts can be intentionally suppressed by specifically prompting the LLM to do that. While our manual assessment suggests that suppressing the model's internal knowledge is indeed feasible, which can be seen in the opening example with Václav Havel's award (Figure 2), the definitive confirmation of this hypothesis would require additional research.

In terms of the methodological approach, we made a simplifying assumption that it is rare for information sources to contain internal contradictions. Based on this assumption, it is sufficient for our method to confirm a statement if at least one paragraph (text chunk) supports it. What would happen if we took the opposite approach and evaluated a document as supporting a given statement only if none of the paragraphs in that document disputed the statement (and at least one statement supported it)? In this case, our experience led us to believe that we would obtain much lower recall because only a slight uncertainty in one paragraph could cause the large language model to assess it as a contradiction due to excessive caution or due to a potential hallucination, ultimately leading to the entire document being disregarded. Furthermore, we believe that there are very few internally contradictory sources, making such cases rare.

One possibility for further improvement of both precision and recall simultaneously is adapting a method similar to the one proposed by Ding et al. (2024), who devised a new method for improving the recall of large language models in the relation extraction task by combining the large language model with an evaluation-filtering model based on the transformer architecture that generates candidate relational entity pairs with higher recall.

In the future, we intend to improve the implementation of the verification workflow on Wikidata. To make the process effective, we would like to make a narrower selection of data to be verified because most Wikidata entries are believed to be true and factual. For example, we could focus on the claims with multiple values that have one or more mandatory required qualifier constraint violations, which are presented as possibly conflicting.

The performance of the web search lookup could be improved by factoring in a disambiguation element. In the current version, the web search query is constructed only from the given statement. In future work, we plan to experiment with query expansion by adding contextual information extracted from connected Wikidata statements.

The evaluation was performed on the BioRED dataset, which was expanded with negative statements, and additionally on the SNLI dataset. The advantages of this approach were that we evaluated against datasets used in related work, and we could obtain estimates of predictive performance without the need for manual checking, which is typically used in knowledge graph statement validation. However, as noted above, this approach may not fully represent performance on Wikidata, and therefore an important next step is to perform the evaluation directly on Wikidata or a similar knowledge graph, such as DBpedia (Bizer et al., 2009).

**CRediT authorship contribution statement**

**Daniel Adam**: Writing – original draft, Writing – review and editing, Investigation, Methodology, Data curation, Resources, Software, Validation, Visualization. **Tomáš Kliegr**: Writing – original draft, Writing – review and editing, Conceptualization, Methodology, Supervision.

**Data availability**

The code and data are available at https://github.com/danieladam2001/.

**Acknowledgements**

The authors would like to thank the Faculty of Informatics and Statistics, VSE, which supported this research through long-term institutional support of research activities.